# Handling Uncertainty during Plan Recognition in Task-Oriented Consultation Systems

Bhavani Raskutti
Computer Science Department
Monash University
Clayton, VICTORIA 3168
AUSTRALIA

Ingrid Zukerman
Computer Science Department
Monash University
Clayton, VICTORIA 3168
AUSTRALIA

## Abstract

During interactions with human consultants, people are used to providing partial and/or inaccurate information, and still be understood and assisted. We attempt to emulate this capability of human consultants in computer consultation systems. In this paper, we present a mechanism for handling uncertainty in plan recognition during task-oriented consultations. The uncertainty arises while choosing an appropriate interpretation of a user's statements among many possible interpretations. Our mechanism handles this uncertainty by using probability theory to assess the probabilities of the interpretations, and complements this assessment by taking into account the *information content* of the interpretations. The information content of an interpretation is a measure of how well defined an interpretation is in terms of the actions to be performed on the basis of the interpretation. This measure is used to guide the inference process towards interpretations with a higher information content. The information content of an interpretation depends on the specificity and the strength of the inferences in it, where the strength of an inference depends on the reliability of the information on which the inference is based. Our mechanism has been developed for use in task-oriented consultation systems. The domain that we have chosen for exploration is that of a travel agency.

## 1 INTRODUCTION

During task-oriented consultations, a consultant needs to infer a user's requirements from his/her statements in order to provide assistance. To this effect, the consultant needs to interpret a user's statements correctly. However, this task is hindered by the fact that people often provide partial and/or inaccurate information. This requires the consultant to fill in the missing information by using different information sources, such as knowledge of discourse coherence, domain knowledge and knowledge about the user. However, these information sources are not fully reliable, requiring the consultant to draw inferences which are inherently uncertain. As a result, the statements issued by the user may be interpreted in more than one way. Hence, the consultant needs to evaluate the possible interpretations and select the most probable one. In this paper, we present a mechanism for handling the uncertainty arising from the lack of reliability of the various information sources used by the consultant, and for discriminating between multiple interpretations of a user's statements.

An interpretation of a user's statements consists of a sequence of plans that the user proposes to carry out, and a plan consists of an action with a number of parameters defining the action. For instance, in the travel domain, the proposal to fly from Melbourne to Sydney on December 1st, 1990, is a plan, where *flying* is the action, and the parameters *origin, destination* and *departure date* are instantiated.

A number of researchers have used plan recognition as a means to response generation during consultation (Grosz 1977, Allen and Perrault 1980, Sidner and Israel 1981, Carberry 1983, Litman and Allen 1987, Pollack 1990). However, the models of plan recognition developed by these researchers cope only with a single interpretation of a user's actions or utterances. Carberry (1990) addresses the problem of multiple interpretations by using default inferences, and by applying Dempster-Shafer theory of evidence (Section 9.1, Pearl 1988) to compute plausibility factors of alternate hypotheses. However, in domains such as travel, where the default assumptions are weak, this approach alone does not cope with the problem of multiple interpretations. Kautz and Allen (1986) use circumscription to generate all possible interpretations during story understanding. However, since all possibilities are constructed, and the search is limited only by the struc-ture of the plan hierarchy, this approach can be very expensive.

Our mechanism applies Bayesian theory of probability in order to address the problem of multiple interpretations which result from uncertain information sources. Probability theory in the form of Bayes Belief Networks (Pearl 1988) was applied by Goldman and Charniak (1989) to model the difference between the effect of objects on plan recognition in stories and in 'real life'. In an earlier research, Goldman and Charniak (1988) combined probability theory with Assumption-based TMS (de Kleer 1986) for plan inference during story understanding. Our mechanism expands on this earlier research with respect to plan inference during task-oriented consultations. We have observed that in these consultations, the user generally provides sufficient information to enable the consultant to help him/her achieve his/her goals. Hence, the better defined the plans in an interpretation, the more likely it is that this is the interpretation intended by the user. Based on this observation, we augment probabilistic reasoning by means of information content considerations, in order to narrow down the number of interpretations produced by the system from a user's statements. The information content of an interpretation is a measure of how well defined an interpretation is in terms of the actions to be performed on the basis of the interpretation. It depends on the specificity and the strength of the inferences in it, where the strength of an inference depends on the reliability of the information source on which it is based.

The use of information content considerations is valid only in intended plan recognition, such as the one occurring in cooperative interactions. In *keyhole* recognition of plans, where the observer infers an agent's plans and goals by unobtrusively observing the agent (Schmidt, Sridharan and Goodson 1978), we cannot assume the same strong desire for communication and hence cannot use the information content of an interpretation to assess its probability.

The subsequent sections describe the inference mechanism with particular reference to the means used for handling the uncertainties arising during the inference process. The actual procedures which draw the inferences are discussed in [Raskutti and Zukerman 1991]. We make use of the following dialogue excerpt to illustrate our approach:

Traveler: "I want to go to Sydney the day after tomorrow. I am going to Hawaii on the 11:00 am flight. By the way, I'll be leaving from Adelaide."

## 2    THE INFERENCE MECHANISM

The inference mechanism operates on input provided by a Natural Language Interface (NLI), which consists of predicates, such as FLY and LEAVE, and meta predicates that indicate the modality of the statements, such as CAN and MUST. Based on this input, it generates the intended plans of the user. The inference mechanism consists of three processes: (1) Direct inference, (2) Indirect Inference, and (3) Evaluation of interpretations. In these processes, the uncertainties arising due to partial and possibly unreliable information are handled by using Bayesian theory of probability combined with information content considerations.

The direct inference process generates a set of possible interpretations from the input provided by the NLI, using definitions of domain actions and coherence considerations. During this stage, there are uncertainties arising due to the many possible interpretations of a statement as well as due to the many possible relations between the interpretation of a new statement and the previous discourse. For instance, in the above dialogue excerpt, there is uncertainty as to which trip the departure location 'Adelaide' refers to. This uncertainty is handled by computing the probability of an interpretation of a piece of discourse in terms of the probability of an interpretation of each new statement (Section 3.1), the probability of an interpretation of the previous discourse, and the probability of a relation between the two (Section 3.2). The calculated probabilities are used to prune the set of interpretations (Section 3.3).

The interpretations generated by the direct inference process are usually incomplete. The unspecified details are filled in by indirect inferences based on other information sources, such as domain knowledge and world knowledge. All these sources are not equally reliable and hence the strength of these inferences are not the same. For instance, the desired mode of transport between Sydney and Hawaii may be inferred by taking into consideration typical assumptions about the domain. This type of inference is stronger than an inference based on general 'world knowledge', but weaker than a direct inference. During the indirect inference process, the strength of an inference is used to prevent a weaker inference from refuting the results of a previous stronger inference (Section 4).

A measure of information content is used both after the direct inference process and during the indirect inference process to determine whether the inference process should be continued. This measure is also used after both the direct and the indirect inference process to increase the probability of the interpretations with a high information content and to decrease the probability of those with a low information content (Section 5). The updated probabilities are used to prune the set of interpretations by dropping the interpretations with a low probability (Section 3.3). Thus, the evaluation process prefers those interpretations that are well-defined, i.e., those that were completely specified by the user or those in which the gaps left by the user could be filled in by using the system's knowledge.

## 3  THE PROBABILITY OF AN INTERPRETATION OF THE DISCOURSE

The probability of an interpretation of a set of statements is a measure of how likely it is that the speaker intended this interpretation when s/he uttered the statements in question. Assume that our discourse $S'$ is composed of the previous discourse $S$ and a new statement $s$, where $s$ stands for the input returned by the NLI and consists of a predicate and a possibly *nil* meta predicate. Further, assume that $S$ has a set of possible interpretations $\{I_j\}$, and $s$ has a set of possible interpretations $\{i_k\}$. In addition, assume that $\{R_m\}$ is the set of possible relations between an interpretation $I_j$ of $S$ and an interpretation $i_k$ of $s$. Each relation $R_m$ is one of the possible discourse relations between a statement and previous discourse, namely, Elaboration, Introduction and Correction. In addition, the relations Elaboration and Correction must also refer to a topic or plan that is being elaborated or corrected. Finally, let $D$ be the set of apriori domain knowledge of the listener including the plan-base and the rule-base that are used to generate the set of interpretations. Thus, the probability that the speaker meant an interpretation $I_{jkm}$, consisting of $I_j$ and $i_k$ and the relation $R_m$ between $I_j$ and $i_k$, when s/he said $S'$ is $P(I_{jkm}|S', D)$. However, since we are dealing with a closed system where the domain knowledge remains constant throughout the interaction, we can omit $D$ from our calculations, and focus only on $P(I_{jkm}|S')$. This probability is calculated as follows:

$$P(I_{jkm}|S') = P(I_j, i_k, R_m | S, s)$$

According to Bayes Rule for conditional probability:

$$P(I_{jkm}|S') = \frac{P(I_j, i_k, R_m, S, s)}{P(S, s)}$$

Using Bayes Rule for conditional probability, the numerator can be rewritten as follows:

$$P(I_j, i_k, R_m, S, s) =$$
$$P(S|s, i_k, I_j, R_m) * P(s|i_k, I_j, R_m) *$$
$$P(R_m|i_k, I_j) * P(I_j|i_k) * P(i_k)$$

- $P(S|s, i_k, I_j, R_m) = P(S|I_j)$, since $S$ is conditionally independent of $s$, $i_k$ and $R_m$, given $I_j$; $P(S|I_j)$ indicates the probability that a user would utter the statements in discourse $S$ when s/he wanted to mean $I_j$.

- $P(s|i_k, I_j, R_m) = P(s|i_k)$, since $s$ is conditionally independent of $I_j$ and $R_m$, given $i_k$; $P(s|i_k)$ indicates the probability that a user would utter the statement $s$ when s/he wanted to mean $i_k$.

- $P(I_j|i_k) = P(I_j)$, since $I_j$ is the interpretation of statements before $s$ and is independent of $i_k$.

Using the above results, the expression for $P(I_{jkm}|S')$ may be rewritten as follows:

$$P(I_{jkm}|S') =$$
$$\frac{P(S|I_j) P(s|i_k) P(R_m|i_k, I_j) P(I_j) P(i_k)}{P(S, s)}$$

The application of Bayes Rule to the first two terms of the numerator yields:

$$P(I_{jkm}|S') =$$
$$\left( \frac{P(I_j|S) P(S)}{P(I_j)} \right) * \left( \frac{P(i_k|s) P(s)}{P(i_k)} \right) *$$
$$\left( \frac{P(R_m|i_k, I_j) P(I_j) P(i_k)}{P(S, s)} \right)$$

The terms $P(s)$, $P(S)$ and $P(S, s)$ represent how often the statement $s$, the discourse $S$ and the discourse $S'$ consisting of $S$ and $s$ are ever uttered, regardless of the purpose for which they are uttered. These probabilities, in general, can be very difficult to estimate. However, since they are independent of the probabilities of the interpretations, and since we are interested in comparisons of probabilities and not in absolute values, we can define a normalizing constant $\alpha$, $\alpha = \frac{P(S) \times P(s)}{P(S, s)}$. Hence,

$$P(I_{jkm}|S') = \alpha P(I_j|S) P(i_k|s) P(R_m|i_k, I_j)$$

$P(I_j|S)$ is the probability of the interpretation $I_j$ after processing the discourse $S$. It is the result of iteratively applying the process described here with respect to the statements in discourse $S$. $P(i_k|s)$ is the probability of an interpretation $i_k$ of a user's statement $s$ and it is computed as described in Section 3.1. $P(R_m|i_k, I_j)$ links up the new statement's interpretation to the interpretation of the earlier discourse by means of relation $R_m$. It is determined as described in Section 3.2. The probabilities of the new interpretations, $P(I_{jkm}|S')$, for all the combinations of $\{I_j\}$, $\{i_k\}$ and $\{R_m\}$, are used to prune the set of interpretations.

### 3.1  PROBABILITY OF AN INTERPRETATION OF A STATEMENT

The input processed by the inference mechanism is a parsed version of the original statements issued by the user, and consists of a predicate and a possibly *nil* meta predicate for each statement. During the direct inference process, interpretations of an input predicate are determined by using an operator library (Fikes and Nilsson 1971) and plan inference rules (Allen and Perrault 1980). The operator library defines the basic actions in the domain. Each operator definition in the library consists of the preconditions, effects and body of an action, where the body defines the composition of the action. The plan inference rules state that all those operators that have the input predicate in their precondition, effect or body must be chosen as possible interpretations of the predicate.

The probability of an interpretation $i_k$ is determined as follows: all the interpretations generated by the same rule, e.g., a body rule, are assigned the same probability; and as more possibilities are generated, each individual one is considered less probable. However, the total probability mass allocated to all the interpretations resulting from the application of a rule depends on two factors: (1) the meta predicate returned by the NLI, and (2) whether other rules gave rise to the inference of any interpretations.

The presence of a meta predicate, such as WANT, enables us to resolve ambiguity when the input predicate appears in both the precondition and the effect of an operator. For instance, if a user's request about BEing at a place is expressed as "I want to be ... " or "I can be ...," but the NLI returns it as BE(...), this predicate can refer either to the precondition or the effect of an operator with equal probability. By taking into consideration the presence of a meta predicate, we increase the bias towards the appropriate inference rule, e.g., if the user had said "I can be ...," then the probability mass of the precondition rule would have been increased. This scheme is implemented by determining in advance the manner in which each of the meta predicates that may be returned by the NLI affects the probability mass assigned to each rule.

A rule which does not match an input predicate fails to infer any interpretations. If one or more rules fail to infer interpretations, the probability mass allocated to the rules that were successful is modified by allocating the apriori mass of the unsuccessful rule proportionally between the successful ones. For instance, if only the effect and precondition rules yield interpretations, then the probability mass allocated to the body rule is 0. This results in the apriori allocation of the body rule being split up proportionally between the effect and precondition rules.

These considerations for determining the probability of an interpretation are domain independent, since the prior probabilities of all the interpretations of a predicate are considered equal, and they are modified using only meta predicates. At this juncture, if domain knowledge is available, it can be used to modify the apriori probabilities of the interpretations of a predicate.

## 3.2 PROBABILITY OF A RELATION

The possible discourse relations considered by the inference mechanism are as follows: Elaboration, Correction, Digression and Introduction. Digression is a special case of Elaboration, where the probability of elaborating on the last topic is considerably reduced.

Hence, the relation $R_m$ can only be one of the other three. In addition, in the case of Elaboration and Correction, $R_m$ also includes the topic that is being referred to. The possible relations and the number of topics that can be referred to lead to a large number of elements in the set $\{R_m\}$. This set is constrained by assigning probabilities to the inferred relations so that normal patterns of discourse are preferred. The following considerations are used while determining the probability of an inferred relation:

1. When a new statement can be interpreted as elaborating on two or more topics discussed earlier, the elaboration of the last referenced topic is preferred to the elaboration of earlier topics. Further, the probability of an earlier topic being referred to falls exponentially as its distance from the current statement increases.

2. A new statement can always be interpreted as introducing a new topic of discussion. However, if the new statement can also be interpreted as elaborating on an earlier topic, the probability of introduction is considerably reduced. This is achieved by assigning a constant and low probability to the introduction relation. If a highly probable elaboration relation is possible, then the relative magnitude of the probability of introduction is reduced.

3. If there are cue words that indicate a particular relation or that point to a particular previous topic, the probability of this relation or this topic increases. For instance, in the statement "*By the way*, I'll be leaving from Adelaide," from the dialogue excerpt presented in Section 1, the italicized cue phrase indicates a digression. Hence, the preferred interpretation is that 'Adelaide' refers to the trip that was mentioned earlier in the discourse, namely the Sydney trip.

4. Correction and digression are never inferred unless the NLI provides evidence to this effect by means of cue phrases, such as "on second thought" and "by the way," respectively.

The above considerations are domain independent, since the probability of a specific relation $R_m$ is determined by the cue words returned by the NLI and by the contents of the interpretations $i_k$ and $I_j$. Domain knowledge of distances is not taken into consideration while calculating the probability of the relation $R_m$. However, the probability of going to Hawaii on the way between Sydney and Melbourne is definitely lower than the probability of going to Sydney on the way between Melbourne and Hawaii. If domain knowledge is available, it should be incorporated into the system to update the probabilities obtained using the domain independent considerations discussed above.

## 3.3 PRUNING THE SET OF INTERPRETATIONS USING THEIR PROBABILITIES

After determining the probabilities of a set of interpretations, the probabilities are normalized. The normalized probabilities are used to prune the set of interpretations by dropping all those interpretations whose probabilities fall below a relative rejection threshold. In principle, the improbable interpretations could be maintained and revisited, if necessary. But, empirically, this has not been necessary since the intended interpretation has been found among the retained interpretations. All the interpretations $I$ that are retained have a probability that satisfies the following condition:

$$\frac{P_I}{P_{max}} \geq Threshold$$

where $P_{max}$ is the maximum of all the normalized probabilities, $P_I$ is the probability of a retained interpretation, and $Threshold$ is a number in [0,1] range. This calculation ensures that interpretations with a low probability relative to the most probable interpretation are dropped. For example, with 0.5 as the value for $Threshold$, an interpretation with probability 0.3 is dropped if there is another interpretation with probability 0.7. At the same time, if there is a situation where there are three interpretations with probability 0.4, 0.3 and 0.3, then all three are retained.

By judicious choice of a value for $Threshold$, the system can be tailored to consider more or less possibilities. For instance, currently we have two thresholds. The first one used during the direct inference process is 0.5 and was chosen in line with the probabilities assigned to the discourse relations, so that plausible interpretations are not discarded. The second threshold, which is used to prune the interpretations after the indirect inference process, is 0.7, and was chosen so that fewer possibilities are considered as additional information is brought to bear.

## 4 STRENGTH OF INFERENCES

We postulate that the strength of an inference is directly proportional to the reliability of the information source that is used as the basis for this inference. Hence, we categorize different information sources that are used to draw inferences and list them below in decreasing order of reliability.

1. *User's Statements* – Direct inferences from what is explicitly stated. While these inferences can be presumed correct, there is still a degree of uncertainty in relating a new statement to the earlier ones due to the different discourse relations possible.

2. *Domain Knowledge* – Indirect inferences that are derived by using the system's beliefs about the user's domain knowledge. A typical example is the inference of the arrival time at the destination once the departure time is known. Such an inference is useful when there are multiple legs in a proposed journey, requiring that departure times at subsequent locations be inferred.

3. *Domain Assumptions* – Indirect inferences that are derived by assuming what is normal in the domain. For example, when no details about the mode of travel are specified, it is possible to derive this information from the usual mode of transport between two places.

4. *User Model* – Indirect inferences that are made on the basis of the system's model of the user. The user model may be a default model describing a typical user, or it may be more specific. In the context of a travel agency, we have adopted a default model based on the assumption that typically, in a travel agency, the information provider cannot form an extensive user model.

5. *Common-Sense* – Indirect inferences that are derived by assuming normal behavior or common notions outside the domain of interest. Typically, such notions are used when we postulate return journeys based on the assumption that people usually do not move from their residence.

The inference types are assigned a strength in the (0,1] range, and this strength is used to calculate the information content of a parameter and also to determine whether a particular parameter should be revised by a new inference during the indirect inference process. The inferences derived from the user's statements have a strength of 1 and all other inference types have a progressively decreasing strength according to the reliability of their source of information. Undefined parameters are assigned a minimum strength. This assignment enables us to distinguish between parameters that are defined inexactly by the user and parameters that are left undefined, and assign less information content to undefined parameters.

In the process of deriving indirect inferences, we emulate one aspect of human behavior whereby once a conclusion is accepted with a particular degree of confidence, people consider it to be certain when drawing subsequent conclusions (Gettys, Kelly and Peterson 1982). Thus, like Carberry (1990), we do not compound the uncertainty in chains of inferences. This approach is different from that used for the computation of confidence factors (CFs) in MYCIN, where the CF of a parameter $P$ is computed by taking into account the CFs of the parameters $P_1, P_2, ..., P_n$ that are used for calculating $P$ (Shortliffe and Buchanan 1975). Our approach is implemented by tagging each parameter, $P$, in the plans in each interpretation with the type of inference that gave rise to the value of $P$, without taking into consideration the inference types of $P_1, P_2, ..., P_n$. Since the type of inference indicates the strength of the inference rule, it represents the conditional probability of $P$, given $P_1, P_2, ..., P_n$.

During the indirect inference process, the strength of the inference in a parameter's tag is compared with the strength of a new inference, before updating the parameter with the new inference. If the strength of the new inference is the same or higher than the strength of the inference type in the tag, the new inference replaces the old one. Otherwise, the old inference is retained. In this manner, a weaker inference is prevented from refuting the results obtained from a stronger inference.

Thus, during the indirect inference process, the gaps left in the interpretations generated by the direct inference process are filled in using the most reliable indirect inference, and unreliable inferences are not maintained. In principle, it is necessary to generate a new interpretation for each new value of a parameter, and process all the generated interpretations. However, this can lead to an exponential explosion of possibilities, and our decision to consider only the strongest inference during the indirect inference process is a trade-off between the benefits of exploring all possibilities and the resource limitations for doing so (Horvitz 1989).

# 5  INFORMATION CONTENT AND ITS USE

The information content of an interpretation is a measure of the extent of its definition. After both direct and indirect inferences, this measure is used both to determine if the processing is to be continued as well as to modify the probabilities of the interpretations so that the probabilities of interpretations with a higher information content are increased. The modified probabilities are used to prune the set of interpretations as described in Section 3.3.

## 5.1  THE INFORMATION CONTENT OF AN INTERPRETATION

We define the information content of an interpretation as the sum of the information content of all the plans in the interpretation, and the information content of a plan as the sum of the information content of all the parameters that are necessary for the definition of the plan. The information content of a parameter, in turn, depends on two factors: (1) its specificity, which is defined as the reciprocal of the number of possible values assigned to this parameter, and (2) its strength, which depends on the source of information from which this parameter was obtained. That is, both a parameter with multiple values assigned to it and a parameter derived from an unreliable source of information are deemed to have a low information content.

We borrow from Information Theory (Shannon 1948), to define the information content of a parameter $p$, $IC(p)$, as follows:

$$IC(p) = \log_2 \frac{S(p)}{N(p)}$$

where $N(p)$ is the number of possible values assigned to $p$, and $S(p)$ is the strength associated with $p$. The strength of the parameter is the strength of the inference type that was used to derive the value of the parameter (Section 4.2).

According to this formula, undefined parameters have the least information content, since they can take on all the possible values in the domain, and parameters inferred exactly from a reliable source, such as a direct inference from a user's statement, have a maximum information content. For instance, if we have directly inferred that the departure date for a trip is between the 9th and the 15th of May, 1991, then the information content of this parameter is $log_2 1/7$. This measure is additive over multiple plans and it ranges over the negative values, with a maximum information content of 0 when all the parameters which are necessary for the definition of a plan are exactly defined. Thus, the information content of an interpretation $I$, $IC(I)$, is:

$$IC(I) = \sum_{\{\text{plans } P_j \text{ in } I\}} \sum_{\{\text{parameters } p_i \text{ in } P_j\}} \log_2 \frac{S(p_i)}{N(p_i)}$$

## 5.2  CHECKING COMPLETION

After the direct inference process and during the indirect inference process, the information content measure is used to determine if the processing should be continued. Processing is stopped if at least one complete interpretation, i.e., an interpretation with zero information content is determined. Processing of an interpretation is also stopped if no new inferences can be drawn on the basis of the existing evidence, i.e., the information content of an interpretation cannot be increased further. Thus, the information content measure is used as an *informative stopping rule* (Berger and Berry 1988) to determine if the processing should be stopped.

## 5.3  UPDATING PROBABILITIES

The information content measure, which ranges over the negative values, is mapped to the [0,1] range and then used to update the probability of an interpretation, $P(I)$. This is performed by means of the following formula:

$$P(I) \leftarrow P(I) \left( 1 - \frac{IC(I)}{ICNORM} \right)$$

$ICNORM$ is currently defined to be the minimum possible information content in the domain. In our

restricted domain, where the number of possible destinations and origins is low, this choice of $ICNORM$ has a large impact on the probability. However, in a realistic domain, where there is a greater degree of freedom in terms of the values that can be assigned to the parameters, another definition of $ICNORM$ may be preferred. Currently, we are experimenting with $ICNORM$ as the sum of the information content of all the interpretations, and use this definition to update the probabilities when there are multiple interpretations. The update of probabilities using information content is valid only in cooperative information-seeking interactions, such as those occurring at a travel agency, where the user wants his/her intentions to be understood by the listener and hence, interpretations with higher information content are more probable.

## 6   EXAMPLES

Our system has been implemented to understand discourses in travel domain. The language used for the implementation is Franz Lisp. Our system has a rule-base containing twelve rules and a plan-base containing seven plan operators. The input to the system is in the form of predicates, and the system produces output in the form of possible interpretations consisting of one or more plans that the user proposes to carry out. To illustrate the inference process, we consider two plausible dialogue excerpts at a Melbourne travel agency.

### EXAMPLE 1

Traveler: "I want to go to Sydney the day after tomorrow. I am going to Hawaii on the 11:00 am flight. By the way, I'll be leaving from Adelaide."

This chunk of statements issued by a traveler is returned by the NLI as the following four predicates, where the last two predicates are due to the third sentence:

(1) GO (departure_date = *day after tomorrow*,
     destination = *Sydney*)
(2) FLY (departure_time = *11:00 am*,
     destination = *Hawaii*)
(3) DIGRESS
(4) LEAVE (origin = *Adelaide*)

The first two domain predicates give rise to an interpretation composed of two plans: (a) to go to *Sydney* and (b) to fly to *Hawaii* at *11:00 am*. The DIGRESS discourse relation in predicate (3) indicates that a plan which precedes the plan currently in focus is likely to be the topic of discussion for the forthcoming predicate. Thus, with the fourth predicate, we have two possible interpretations: $I_1$ — it elaborates plan (a), or $I_2$ — it elaborates plan (b). $I_1$ has a higher probability due to the presence of the DIGRESS discourse relation.

The information content measure rates both these interpretations equally. Hence, both interpretations are retained after the direct inference process. During the indirect inference process, we consider the case where the temporal order of plans is assumed to be the same as the order of presentation during the discourse. Other possible temporal orders are considered and processed in [Raskutti and Zukerman 1991]. The use of indirect inference rules coupled with the assumption for temporal order of plans gives rise to two scenarios:

$I_1$ : $Adelaide \rightarrow Sydney \rightarrow Hawaii$

$I_2$ : $Melbourne \rightarrow Sydney, Adelaide \rightarrow Hawaii$

$I_2$ has less information content, since its parameters cannot be inferred due to the need to postulate an additional intervening plan to take the user from *Sydney* to *Adelaide*. Hence its probability is correspondingly decreased and $I_1$ is chosen as the best interpretation.

### EXAMPLE 2

Traveler: "I want to go to Sydney the day after tomorrow. From Sydney I'll be going to Hawaii on the 11:00 am flight. I'll be leaving from Adelaide."

This chunk of statements issued by a traveler is returned by the NLI as the following three predicates.

(1) GO (departure_date = *day after tomorrow*,
     destination = *Sydney*)
(2) FLY (departure_time = *11:00 am*,
     origin = *Sydney*
     destination = *Hawaii*)
(3) LEAVE (origin = *Adelaide*)

The first two domain predicates give rise to an interpretation composed of two plans: (a) to go to *Sydney* and (b) to fly to *Hawaii* from *Sydney* at *11:00 am*. With the third predicate, we have two possible interpretations: $I_1$ — it elaborates plan (a), or $I_2$ — it introduces a new plan (c) to go from Adelaide. The probability of $I_1$ is higher since elaboration is preferred to introduction. However, since the elaboration is that of a plan discussed before, $I_2$ is retained as a possibility. Thus after the direct inference process, we have two possibilities:

$I_1$ : $Adelaide \rightarrow Sydney,$
     $Sydney \rightarrow Hawaii$

$I_2$ : $?x \rightarrow Sydney,$
     $Sydney \rightarrow Hawaii,$
     $Adelaide \rightarrow ?y$

$I_1$ has higher information content since the origins and destinations of the two proposed trips are known. This coupled with the previous lower probability of $I_2$ ensures that $I_1$ is the only possibility carried over to the indirect inference process. Thus, during the indirect inference process, $I_1$ is completed to yield the same interpretation as the one in the first example.

# 7  CONCLUSIONS

In this paper, we have described a means for handling uncertainty during plan recognition in task-oriented consultation systems by using Bayesian probability theory augmented by an information content measure. We have used our system on five simple discourse samples similar to the one discussed in Section 6. In each case, the system chooses the same interpretation that people choose, indicating that our tenet of linking specificity and strength of inference to the probability of an interpretation can be valuable in handling real conversations in cooperative interactions. Finally, by modifying the definition of the information content measure to suit different domains, our method may be used for interpreting user's statements in general, as well as in the area of multi-media document retrieval.

## Acknowledgments

This research was supported in part by grant Y90/03/22 from the Australian Telecommunications and Electronics Research Board. We thank Prof. J. Roach and D. Sanford from the Virginia Polytechnic Institute and State University for allowing us to use their transcripts of telephone conversations at travel agencies. We also thank Wilson Wen from Telecom Research Laboratories for his advice on probability theory.